\documentclass[runningheads]{llncs}

\usepackage{stackrel}
\usepackage{multirow}
\usepackage{makecell}
\usepackage{siunitx}
\usepackage{graphicx}

\usepackage[colorlinks=true, urlcolor=blue, citecolor=blue, filecolor=blue, anchorcolor=blue, linkcolor=blue]{hyperref}
\usepackage{siunitx, graphicx, multirow, booktabs}
\usepackage{url}
\begin{document}

\title{RuBioRoBERTa: a pre-trained biomedical language model for Russian language biomedical text mining}


\author{
Alexander Yalunin \and
Alexander Nesterov \and 
Dmitriy Umerenkov
}

\authorrunning{A. Yalunin et al.}
\titlerunning{RuBioRoBERTa: a pre-trained biomedical language model}

\institute{Sberbank Artificial Intelligence Laboratory, Moscow, Russia \\
\email{ale.yalunin@gmail.com, ainesterov@sberbank.ru, d.umerenkov@gmail.com} \\
}

\maketitle % typeset the header of the contribution

\begin{abstract}
%The abstract should briefly summarize the contents of the paper in
%15--250 words.

This paper presents several BERT-based models\footnote{\url{https://github.com/sberbank-ai-lab/RuBioRoBERTa}} for Russian language biomedical text mining (RuBioBERT, RuBioRoBERTa). The models are pre-trained on a corpus of freely available texts in the Russian biomedical domain. With this pre-training, our models demonstrate state-of-the-art results on RuMedBench – Russian medical language understanding benchmark that covers a diverse set of tasks, including text classification, question answering, natural language inference, and named entity recognition. 
%The models and the code are freely available \footnote{\url{https://github.com/sberbank-ai-lab/RuBioRoBERTa}}.

\end{abstract}

\section{Motivation}

The transition to electronic health records led to an overwhelming number of medical data being stored digitally. With the increasing availability of medical texts and the rapid progress of text processing models, biomedical text mining is a rapidly expanding field. Different players from startups to medical institutions build applied solutions to solve medical tasks using biomedical texts. The vast majority of these solutions use the transformer neural network models pre-trained on large text corpora. Such pre-training is computationally expensive and requires the assembly of a corresponding dataset. Thanks to the community effort, we now have access to pre-trained models for all languages. For this work, we used transformer networks pre-trained on Russian language texts from SberDevices\footnote{\url{https://github.com/sberbank-ai/model-zoo}} and DeepPavlov \cite{kuratov2019adaptation}.

A series of models with novel BERT architecture, e.g., ClinicalBERT \cite{alsentzer-etal-2019-publicly}, BlueBERT \cite{peng2019transfer}, successfully applied to a range of healthcare and biomedical domain tasks. In BioBERT \cite{lee2020biobert} authors showed that additional pre-training of transformer models on biomedical domain texts leads to noticeable improvements over a wide range of corresponding domain tasks. We do the same thing for the Russian language: we pre-train three transformer models on a large corpus of Russian medical texts and evaluate on the recently released medical benchmark RuMedBench \cite{blinov2022rumedbench}. With this pre-training, our models demonstrate state-of-the-art results. We hope that the presented models will be used as baselines for a wide array of applied biomedical models.

%  We pre-trained models of different sizes, with the bigger model achieving better results than the smaller models, at the expense of computational cost.

\section{Setup}

\subsection{Biomedical Corpus for pre-training}
Our created biomedical texts corpus is based  on the public Russian-language scientific articles database CyberLeninka\footnote{\url{https://cyberleninka.ru/}}. All articles presented on this resource are freely distributed with the license CC-BY. For our work, we use articles related to the categories: fundamental medicine, clinical medicine, health sciences, biotechnology, and medical technology. Each article we use contains a title, abstract, and publication text. The final corpus includes 338,000 articles published from 1929 to 2021. The total volume of the corpus consists of 1.2 billion words.

\subsection{Downstream Tasks}
This paper aims to measure language learning abilities in the medical domain. As such, we study downstream performance on a diverse set of benchmarks, including question answering, natural language inference, named-entity recognition, and text classification. Precisely, we measure performance on the RuMedBench \cite{blinov2022rumedbench} medical text classification meta-benchmark, which consists of five downstream tasks.

In RuMedTop3 the task is to predict the three most probable ICD-10 codes related to the deceases based only on raw patient symptoms. In RuMedSymptomRec based on a given text premise, recommend a relevant symptom. In RuMedDaNet given a medical context and a question, give a yes/no answer. The contexts come from diverse fields: therapeutic medicine, pharmacology, biochemistry, human physiology, and anatomy. In RuMedNLI given premise and hypothesis to infer the relation as either entailment, contradiction, or neutral. The premises are extracted from the Past Medical History section in MIMIC-III records, and hypotheses are generated by the clinicians. Finally, in RuMedNER the task is to classify words in drug therapy product reviews.

\subsection{Models}
We take three publicly available checkpoints of BERT \cite{devlin2018bert} and RoBERTa \cite{liu2019roberta} based  models, which were pre-trained on general domain corpora: RuBERT \cite{kuratov2019adaptation}, ruBERT and ruRoBERTa\footnote{\url{https://github.com/sberbank-ai/model-zoo}}. 
The first is an adapted multilingual version of BERT pre-trained by DeepPavlov on Russian Wikipedia. The second and the third are the version of BERT and RoBERTa pre-trained by SberDevices on Russian Wikipedia and Taiga corpus.
We additionally pre-train these models on our biomedical corpus, resulting in bio-versions: RuBERT+Bio, RuBioBERT, and RuBioRoBERTa, respectively. 
Finally, we show the effectiveness of our approach in biomedical text mining by fine-tuning and evaluating them on RuMedBench.

\subsection{Training Details}
We use the HuggingFace \cite{wolf2020transformers} library to obtain models and training scripts. Each model is trained with default hyper-parameters if not stated otherwise.

The BERT-based models have a vocabulary size of 120K subword units, 178 million parameters, and they were pre-trained on 30GB general text corpora. The BERT-based models have a vocabulary size of 50K subword units, 355 million parameters, and they were pre-trained on 250GB general text corpora. 

Each model is pre-trained on the biomedical corpus for 1 epoch with batch size 64 and weight decay of 0.01. The learning rate is warmed up over the ﬁrst 20000 steps to a peak value of 5e-5, and then linearly decays. The whole pre-training took 12 days on three NVIDIA V100 GPUs.

The hyperparameters are identical for all tasks on the fine-tuning stage: we train for 10 epochs with a batch size of 32 and weight decay of 0.01. The learning rate is warmed up over the ﬁrst 30\% of steps to a peak value of 3e-5, and then decays with cosine annealing.

\section{Results}

\begin{table}
\vspace{-4mm}
\centering
\caption{Performance metrics (\%) on the \emph{RuMedBench} test sets.} \label{bl_results}
\begin{tabular}{r c c c c c c} 
\multicolumn{1}{r}{Model} & $\stackrel{Acc/Hit@3}{\mathrm{Top3}}$ & $\stackrel{Acc/Hit@3}{\mathrm{SymptomRec}}$ & $\stackrel{Acc\textcolor{white}{/}}{\mathrm{DaNet}}$ & $\stackrel{Acc\textcolor{white}{/}}{\mathrm{NLI}}$ & $\stackrel{Acc/F1}{\mathrm{NER}}$ & Overall \\
\hline

\multicolumn{1}{r}{Previous best} & 47.45/70.44 & 34.94/52.05 & 71.48 & 77.29 & ~96.47/73.15 & 67.20 \\
\hline
\multicolumn{1}{r}{RuBERT} & 41.24/65.57 & 20.59/34.59 & 68.75 & 77.43 & 96.29/72.57 & 62.32 \\
\multicolumn{1}{r}{ruBERT} & 44.04/69.71 &  23.69/38.09 & 64.06 & 78.06 & 96.59/74.07 &  63.04 \\
\multicolumn{1}{r}{ruRoBERTa} & 45.74/72.14 & 40.92/54.37 & 74.61 & 82.42 & 97.09/77.79 & 70.21 \\
\hline

\multicolumn{1}{r}{RuBERT+Bio} & 45.13/72.26 & 26.92/41.99 & 51.17 & 78.62 & 96.45/74.04 & 61.64 \\
\multicolumn{1}{r}{RuBioBERT} & 43.55/68.86 & 28.94/44.55 & 53.91 & 80.31 & 96.63/75.97 & 62.69  \\
\multicolumn{1}{r}{RuBioRoBERTa} & {\bf 46.72/72.87} & {\bf 44.01/58.95} & {\bf 76.17} & {\bf 82.77} & {\bf 97.19/77.81} & {\bf 71.54} \\

\hline
\multicolumn{1}{r}{Human} & 25.06/48.54 & 7.23/12.53 & 93.36 & 83.26 & 96.09/76.18 & 61.89 \\
\hline
\end{tabular}
\end{table}

As shown in Table \ref{bl_results}, our best model RuBioRoBERTa outperforms the previous state-of-the-art on all tasks and demonstrates the highest overall score. The obvious explanation of why RuBioRoBERTa is the best model among those we present is that it has the largest number of parameters, and it was pre-trained longer on general corpora. Another observation is that "bio" models outperform their correspondent "general" versions in most tasks. It validates our approach in biomedical text mining and additional pre-training.
Like in previous work, our model outperforms assessor-clinicians in the first two tasks. In addition, RuBioRoBERTa now beats human in RuMedNER, i.e., in drug review classification.
The RuMedDaNet is the most challenging one of the proposed tasks, as it requires a deeper understanding of the relation between the context and the hypothesis. We hope that the current lag between the human level and the best model of more than 15\% will be reduced after more advanced and specialized Russian medical models appear.

\section{Qualitative results}

\begin{table}
\vspace{-4mm}
\caption{Prediction samples from ruRoBERTa and RuBioRoBERTa on RuMedBench.} 
\label{qual_table}
\begin{tabular}{p{0.08\linewidth} p{0.20\linewidth} p{0.72\linewidth}} 
Task & Model & Sample \\
\hline

% NER
NER & ruRoBERTa & So they kept us on \textbf{syrups}, like \emph{bronchomax}, and as a result, pneumonia and a hospital. \\
& RuBioRoBERTa & So they kept us on \textbf{syrups}, like \emph{bronchomax}, and as a result, \underline{pneumonia} and a hospital. \\

% TOP3
Top3 & Input, \email{Target} & Pain in the wrist joints,  when lifting weights, rotation. \email{Other arthritis} \\
& ruRoBERTa & Polyarthrosis, Seropositive rheumatoid arthritis, Other rheumatoid arthritis \\
& RuBioRoBERTa & \textbf{Other arthritis}, Other rheumatoid arthritis, Other arthrosis \\

% SYM_REC
Sym & Input, \email{Target} & Pronounced dry cough, increased since the last night, runny nose, headache, weakness, - not measured, chills. \email{Increase in body temperature} \\
& ruRoBERTa & Sore throat, Runny nose, Dry cough \\
& RuBioRoBERTa & Chills, \textbf{Increase in body temperature}, Yellow colored urine \\

% NLI
NLI & Input, \email{Target} & Sentence 1: ACA in XXX with residual left-sided weakness, Sentence 2: Had a stroke. \email{entailment} \\
& ruRoBERTa & neutral \\
& RuBioRoBERTa & entailment \\
 
\hline
\end{tabular}
\end{table}

We conduct a qualitative analysis of models results to explore the sources of improvements. Table \ref{qual_table} gives examples of the predictions from ruRoBERTa and RuBioRoBERTa. For a significant amount of cases, RuBioRoBERTa was able to understand the standard medical abbreviations. In the NLI example, the model needs to know that ACA stands for Acute Cerebrovascular Accident to answer whether the person had a stroke correctly. Also, for several cases, the biomedical pre-training allowed the model to more confidently recognize drug and diseases names like \emph{pneumonia} in the NER example. In the Top3 task, the symptoms listed in the input are not characteristic of any particular rheumatic disease. It was correctly interpreted by RuBioRoBERTa, placing the diagnosis corresponding to nonspecific arthritis \emph{(Other arthritis)} in the first place. In the SymRec task, the domain-specific model also performed better, recommending symptoms not found in the original text, which is not the case for the non-domain-specific model. %Extended examples of models' performance on all the tasks are available in the appendix.

\section{Clinical Relevance}
We believe it is crucial to develop benchmarks and models for languages other than English. The models that we propose can directly support the Russian research community that develops AI medicine methods. Concretely, the models pre-trained in the medical domain that we publish can be used as a starting checkpoint for the textual models used in medicine and healthcare. Also, our models can serve as comparison baselines for works in multilingual/cross-lingual fields that solve medical tasks in different languages.

\section{Conclusion}
This study introduces RuBioRoBERTa, a pre-trained biomedical language model that demonstrates state-of-the-art results on the RuMedBench benchmark. We conduct qualitative analysis to verify our approach in biomedical text mining. We release the source code and the weights of RuBioRoBERTa and RuBioBERT. Our further plan is to increase the volume and quality of the biomedical text corpus, utilize the latest models and techniques in NLP, increase the computational resources and support the development of the RuMedBench benchmark.

%\bibliographystyle{splncs04}
%\bibliography{ref}

\begin{thebibliography}{1}
\providecommand{\url}[1]{\texttt{#1}}
\providecommand{\urlprefix}{URL }
\providecommand{\doi}[1]{https://doi.org/#1}

\bibitem{alsentzer-etal-2019-publicly}
Alsentzer, E., Murphy, J., Boag, W., Weng, W.H., Jin, D., Naumann, T.,
  McDermott, M.: Publicly available clinical {BERT} embeddings. In: Proceedings
  of the 2nd Clinical Natural Language Processing Workshop. pp. 72--78.
  Association for Computational Linguistics, Minneapolis, Minnesota, USA (Jun
  2019)

\bibitem{blinov2022rumedbench}
Blinov, P., Reshetnikova, A., Nesterov, A., Zubkova, G., Kokh, V.: Rumedbench:
  A russian medical language understanding benchmark. arXiv preprint
  arXiv:2201.06499  (2022)

\bibitem{devlin2018bert}
Devlin, J., Chang, M.W., Lee, K., Toutanova, K.: Bert: Pre-training of deep
  bidirectional transformers for language understanding. arXiv preprint
  arXiv:1810.04805  (2018)

\bibitem{kuratov2019adaptation}
Kuratov, Y., Arkhipov, M.: Adaptation of deep bidirectional multilingual
  transformers for russian language. arXiv preprint arXiv:1905.07213  (2019)

\bibitem{lee2020biobert}
Lee, J., Yoon, W., Kim, S., Kim, D., Kim, S., So, C.H., Kang, J.: Biobert: a
  pre-trained biomedical language representation model for biomedical text
  mining. Bioinformatics  \textbf{36}(4),  1234--1240 (2020)

\bibitem{liu2019roberta}
Liu, Y., Ott, M., Goyal, N., Du, J., Joshi, M., Chen, D., Levy, O., Lewis, M.,
  Zettlemoyer, L., Stoyanov, V.: Roberta: A robustly optimized bert pretraining
  approach. arXiv preprint arXiv:1907.11692  (2019)

\bibitem{peng2019transfer}
Peng, Y., Yan, S., Lu, Z.: Transfer learning in biomedical natural language
  processing: an evaluation of bert and elmo on ten benchmarking datasets.
  arXiv preprint arXiv:1906.05474  (2019)

\bibitem{wolf2020transformers}
Wolf, T., Chaumond, J., Debut, L., Sanh, V., Delangue, C., Moi, A., Cistac, P.,
  Funtowicz, M., Davison, J., Shleifer, S., et~al.: Transformers:
  State-of-the-art natural language processing. In: Proceedings of the 2020
  Conference on Empirical Methods in Natural Language Processing: System
  Demonstrations. pp. 38--45 (2020)

\end{thebibliography}

\end{document}